\documentclass[10pt]{article} 

\usepackage[preprint]{rlj}           

\usepackage{amssymb}            
\usepackage{mathtools}          
\usepackage{mathrsfs}           
\usepackage{graphicx}           
\usepackage{subcaption}         
\usepackage[space]{grffile}     
\usepackage{url}                
\usepackage{lipsum}             
\usepackage{thmtools}
\usepackage{thm-restate}
\usepackage{import}
\usepackage{booktabs}       
\usepackage{lmodern}        
\usepackage{bm}
\usepackage{physics}
\usepackage{wrapfig}

\definecolor{wb}{rgb}{0, 0.5, 0}

\DeclareMathOperator*{\argmax}{arg\,max}

\newcommand{\statespace}{\mathcal{S}}
\newcommand{\actionspace}{\mathcal{A}}
\newcommand{\probspace}{\mathscr{P}}

\newcommand{\reals}{\mathbb{R}}
\newcommand{\rewdist}{\mathcal{R}}
\newcommand{\datax}{\mathcal{X}}
\newcommand{\datay}{\mathcal{Y}}
\newcommand{\datac}{\mathcal{C}}
\newcommand{\nngpk}{\kappa}
\newcommand{\ntkk}{\Theta}
\newcommand{\flin}{f_{\text{lin}}}
\newcommand{\tiltheta}{\tilde{\theta}}
\newcommand{\discount}{\gamma}
\newcommand{\ssdist}{\mu}
\newcommand{\transkernel}{P}

\newcommand{\randrew}{R}
\newcommand{\Qpi}{Q^\pi}
\newcommand{\pistar}{\pi^*}

\newcommand{\bonus}{b}

\title{Contextual Similarity Distillation: \\ Ensemble Uncertainties with a Single Model}

\setrunningtitle{Contextual Similarity Distillation: Ensemble Uncertainties with a Single Model}

\author{Moritz A. Zanger\textsuperscript{1}, Pascal R. Van der Vaart\textsuperscript{1}, Wendelin Böhmer \textsuperscript{1}, Matthijs T.J. Spaan \textsuperscript{1}}

\emails{\{m.a.zanger,p.r.vandervaart-1,m.t.j.spaan,j.w.bohmer\}@tudelft.nl}

\affiliations{
$^{1}$\textbf{Delft University of Technology, The Netherlands}\\
}

\contribution{
    This paper introduces contextual similarity distillation, a method for estimating deep ensemble variances efficiently with a single deep neural network, trained with gradient descent, in one forward pass.
    }
    {
    Principled uncertainty estimation remains computationally expensive in many deep learning settings, for example requiring training of an ensemble of models \citep{lakshminarayananSimpleScalablePredictive2017}, complex posterior approximations or samples\citep{izmailov2021bayesian}, or large matrix inversions \citep{rasmussen2006gp}.
    }

\contribution{
    We provide empirical evidence of our approach's practical effectiveness on a variety of visual distribution shift detection tasks and a set of hard exploration problems in deep reinforcement learning. 
        }
    {
    We evaluate several uncertainty quantification methods trained on visual classification datasets (e.g., FashionMNIST) against OOD datasets (e.g., MNIST) using the AUROC, AUPR-IN, and AUPR-OUT metrics. We moreover analyze whether our method provides reliable exploration signals on visual maze exploration tasks from the VizDOOM benchmark suite.
    }


\summary{Uncertainty quantification is a critical aspect of reinforcement learning and deep learning, with numerous applications ranging from efficient exploration and stable offline reinforcement learning to outlier detection in medical diagnostics. The scale of modern neural networks, however, complicates the use of many theoretically well-motivated approaches such as full Bayesian inference. Approximate methods like deep ensembles can provide reliable uncertainty estimates but still remain computationally expensive. In this work, we propose contextual similarity distillation, a novel approach that explicitly estimates the variance of an ensemble of deep neural networks with a single model, without ever learning or evaluating such an ensemble in the first place. Our method builds on the predictable learning dynamics of wide neural networks, governed by the neural tangent kernel, to derive an efficient approximation of the predictive variance of an infinite ensemble. Specifically, we reinterpret the computation of ensemble variance as a supervised regression problem with kernel similarities as regression targets. The resulting model can estimate predictive variance at inference time with a single forward pass, and can make use of unlabeled target-domain data or data augmentations to refine its uncertainty estimates. We empirically validate our method across a variety of out-of-distribution detection benchmarks and sparse-reward reinforcement learning environments. We find that our single-model method performs competitively and sometimes superior to ensemble-based baselines and serves as a reliable signal for efficient exploration. These results, we believe, position contextual similarity distillation as a principled and scalable alternative for uncertainty quantification in reinforcement learning and general deep learning.
}

\begin{document}

\maketitle  

\begin{abstract}
Uncertainty quantification is a critical aspect of reinforcement learning and deep learning, with numerous applications ranging from efficient exploration and stable offline reinforcement learning to outlier detection in medical diagnostics. The scale of modern neural networks, however, complicates the use of many theoretically well-motivated approaches such as full Bayesian inference. Approximate methods like deep ensembles can provide reliable uncertainty estimates but still remain computationally expensive. In this work, we propose contextual similarity distillation, a novel approach that explicitly estimates the variance of an ensemble of deep neural networks with a single model, without ever learning or evaluating such an ensemble in the first place. Our method builds on the predictable learning dynamics of wide neural networks, governed by the neural tangent kernel, to derive an efficient approximation of the predictive variance of an infinite ensemble. Specifically, we reinterpret the computation of ensemble variance as a supervised regression problem with kernel similarities as regression targets. The resulting model can estimate predictive variance at inference time with a single forward pass, and can make use of unlabeled target-domain data or data augmentations to refine its uncertainty estimates. We empirically validate our method across a variety of out-of-distribution detection benchmarks and sparse-reward reinforcement learning environments. We find that our single-model method performs competitively and sometimes superior to ensemble-based baselines and serves as a reliable signal for efficient exploration. These results, we believe, position contextual similarity distillation as a principled and scalable alternative for uncertainty quantification in reinforcement learning and general deep learning.\end{abstract}

\section{Introduction}
With the deployment of increasingly large deep learning systems to real-world applications, efficient uncertainty quantification has become an essential challenge of modern deep learning. Assessing the reliability in predictions is crucial in applications ranging from out-of-distribution (OOD) detection to deep reinforcement learning (RL), where uncertainty estimation is used to drive exploration, stabilize offline learning, increase data efficiency, or to design cautious, safety-aware agents. A necessary condition for designing and deploying such agents is their ability to quantify uncertainty reliably and efficiently. 

Bayesian methods for deep neural networks address this challenge with a solid theoretical footing \citep{goan2020bayesian, pearce2020uncertainty, izmailov2021bayesian} but often require coarse approximations or costly sampling from a complex posterior. To this end, deep ensembles from random initializations \cite{lakshminarayananSimpleScalablePredictive2017, osbandDeepExplorationBootstrapped2016b, qin2022analysis} have emerged as a simple but reliable method for estimating predictive uncertainty in neural networks. While usually more efficient than full Bayesian inference, the computational cost of training several models remains a burden, particularly with increasing parameter spaces. 

In this paper, we introduce contextual similarity distillation (CSD), a novel single-model approach that directly estimates the variance of a random initialization ensemble of deep NNs without ever training or evaluating such an ensemble in the first place. The theoretical motivation for our approach is derived from recent work characterizing the learning dynamics of wide neural networks through the Neural Tangent Kernel \citep[NTK,][]{jacotNeuralTangentKernel2020, leeWideNeuralNetworks2020}. Under some conditions, this setting allows us to describe deep ensembles and in particular their predictive variance by the NTK Gaussian Process \citep[NTK GP,][]{heBayesianDeepEnsembles2020a}, providing an analytical expression for ensemble uncertainties. Although one can in principle solve these analytical expressions explicitly without requiring training of an ensemble of models, these computations quickly become infeasible when considering large models or datasets, as frequently encountered in the field of RL. 

In contrast, we devise a novel method called contextual similarity distillation (CSD) that is amenable to regular training pipelines based on gradient descent and approximates predictive ensemble variance with a single forward pass. We derive our method from the insight that ensemble variance can be obtained as the result of a structured supervised regression problem, where labels correspond to kernel similarities between training points and a test point $x_t$. As a result, one can obtain the predictive variance of a deep ensemble for a known query point $x_t$ by training a single NN on a regression task using gradient descent and a carefully designed label function dependent on $x_t$. We then extend this ``single-query'' approach to work efficiently for arbitrary queries $x_t$ by formulating a contextualized regression model that involves regression tasks with a family of context-dependent label functions. This formulation moreover enables CSD to refine its uncertainty estimates by leveraging unlabeled data, for example from a target domain of interest or from data augmentation techniques, an approach that has proven extraordinarily successful in the field of self-supervised and representation learning \citep{chen2020simple, guo2022byol, caron2021emerging}.

We analyze the practical effectiveness of CSD through an empirical evaluation on a variety of distribution shift detection tasks\citep{van2020uncertainty} using the FashionMNIST, MNIST, KMNIST, and NOTMNIST datasets\citep{xiao2017fashionmnist, deng2012mnist, clanuwat2018kmnist}. We moreover use CSD to generate an exploration signal on sparse-reward reinforcement learning problems from the visual RL benchmark VizDOOM \citep{Kempka2016ViZDoom}. Empirically, CSD consistently achieves competitive and sometimes superior uncertainty estimation to finite deep ensembles and other baseline methods while maintaining lower computational cost. We believe these results establish CSD as a both principled and scalable alternative to ensemble-based uncertainty quantification and exploration methods. 


\section{Background}
For our default framework, we consider a finite Markov Decision Process \citep[MDP,][]{bellmanMarkovianDecisionProcess1957} of the tuple $(\statespace, \actionspace, \rewdist, \discount, P, \ssdist)$, with state space~$\statespace$, action space~$\actionspace$, immediate reward distribution $\rewdist : \statespace \times \actionspace \rightarrow \probspace (\reals)$, discount $\discount \in [0,1]$, transition kernel ${P: \statespace \times \actionspace \rightarrow \probspace(\statespace)}$, and the start state distribution $\ssdist: \probspace(\statespace)$. Here, $\probspace(\mathcal{Z})$ indicates the space of probability distributions over some space $\mathcal{Z}$ and random variables are denoted with uppercase letters. Given a state~$S_t$ at time~$t$, agents choose an action~$A_t$ from a stochastic policy $\pi: \statespace \rightarrow \probspace({\actionspace)}$ and subsequently receives the immediate reward $\randrew_t \sim \rewdist(\cdot|S_t,A_t)$ and observes next state~$S_{t+1} \sim \transkernel(\cdot|S_t,A_t)$. The expected discounted sum of future rewards, conditioned on a particular state $s$ and action $a$ is known as the state-action value and is given by $Q^\pi(s,a) = \mathbb{E}_{\transkernel, \pi}[\sum_{t=0}^\infty \discount ^ t R_t|S_0=s, A_0=a]$. This value function adheres to a temporal consistency condition described by the Bellman equation\ \citep{bellmanMarkovianDecisionProcess1957}
\begin{equation}
    \Qpi(s,a) = \mathbb{E}_{\transkernel, \pi}[\randrew_0 + \discount \Qpi(S_1, A_1)|S_0=s, A_0=a] \,,
\end{equation}
where $\mathbb{E}_{\transkernel, \pi}[\cdot]$ indicates that $S_1$ and $A_1$ are drawn from~$\transkernel$ and~$\pi$ respectively. The expected return of a policy $\pi$ can compactly be expressed through the state-action value and the starting state distribution through
\begin{equation}
    J(\pi) = \mathbb{E}_{S_0 \sim \ssdist, A_0 \sim \pi}[Q^{\pi}(S_0, A_0)] \,.
\end{equation}
The objective of reinforcement learning is to find an optimal policy $\pistar$ that maximizes the above equation $\pistar = \argmax J(\pi)$.

\subsection{Exploration in Reinforcement Learning}
A fundamental challenge in attaining an optimal policy $\pistar$ lies in the exploration-exploitation tradeoff: an agent must decide whether to exploit its current knowledge to maximize returns or whether to explore novel actions in order to discover better strategies. Efficient exploration is particularly crucial in high-dimensional or sparse-reward settings, where naive strategies such as random exploration require prohibitive amounts of interactions. 

A widely used approach to exploration is \textit{optimism in the face of uncertainty} \citep{auer2008near, auerUsingConfidenceBounds2002}, where agents prioritize actions with high epistemic uncertainty in value estimates. In the context of model-free RL, provably efficient algorithms often rely on the construction of an upper confidence bound (UCB) that overestimates the true optimal value $Q^{\pistar}(s,a)$ with high probability\ \citep{jinQlearningProvablyEfficient2018, jinProvablyEfficientReinforcement2019, neustroev2020generalized}. This may be implemented by adding a well-chosen exploration bonus $\bonus(s,a)$ to value estimates according to 
\vspace{-2mm}
\begin{equation} 
Q^{\text{opt}}(s,a) = Q^\pi(s,a) + b(s,a). 
\end{equation}
In small state-action spaces, such bonuses can be derived from count-based concentration inequalities \citep{bellemareUnifyingCountBasedExploration2016, jinProvablyEfficientReinforcement2019}, whereas high-dimensional, continuous domains usually require function approximation, significantly complicating efficient uncertainty estimation \citep{ghavamzadehBayesianReinforcementLearning2015a, osbandDeepExplorationBootstrapped2016b, lakshminarayananSimpleScalablePredictive2017, burda2018exploration}. 


With the widespread use of deep neural networks, deep ensembles \citep{lakshminarayananSimpleScalablePredictive2017} based on random initialization have become a dominant tool for quantifying epistemic uncertainty in high-dimensional continuous spaces \citep{chen2017ucb, osbandDeepExplorationRandomized2019, heBayesianDeepEnsembles2020a}. An informal intuition behind the effectiveness of ensembles is the tendency of randomly initialized NNs to converge to diverse minima in the training loss landscape \citep{fortDeepEnsemblesLoss2020a}, leading to higher prediction diversity for unseen inputs. The variance among ensemble members can then be used to measure the model's uncertainty for a specific input.

\subsection{Neural Tangent Kernel Gaussian Processes}
In order to better understand the properties of deep ensembles and to design better exploration algorithms, an analytical description of deep neural networks and their learning dynamics is desirable. While a general framework remains elusive, significant progress has been made in the field of deep learning theory. In particular, seminal works by \citet{jacotNeuralTangentKernel2020} and \citet{leeWideNeuralNetworks2020} have shown that wide neural networks trained by gradient descent are well-described by their linearized training dynamics and thus predictable. 

For this, let neural networks be parametrized functions $f(x,\theta_t):\reals^n \xrightarrow{}\reals$ and denote training data $\datax = \{x_i \in \reals^n|i\in \{1,...,N_D\}\}$ and training labels $\datay = \{y_i\in \reals|i\in\{1,...,N_D\}\}$. We assume training is performed using gradient descent with infinitesimal step sizes, also referred to as gradient flow. The initialization weights $\theta_0$ are drawn i.i.d. from a normal distribution $\theta_0 \sim \mathcal{N}$, and deep ensembles are formed by training multiple independently initialized neural network functions. We furthermore assume so-called NTK-parametrization, which scales forward and backward passes in proportion to layer widths \citep[see][for details]{jacotNeuralTangentKernel2020, leeWideNeuralNetworks2020}.

A key result by \citet{leeWideNeuralNetworks2020} is that in the limit of infinite layer widths, the training dynamics of deep networks are described exactly by a Taylor expansion around the parameter initialization $\theta_0$. In this setting, the NTK $\ntkk(x,x') : \reals^{n \times n} \xrightarrow{} \reals$, first described by \citet{jacotNeuralTangentKernel2020}, emerges as the defining function governing learning dynamics:
\begin{align}
    \ntkk_0(x,x') = \nabla_\theta f(x,\theta_0)^\top \nabla_\theta f(x',\theta_0).
\end{align}
The NTK can be interpreted as a similarity measure between inputs based on gradient representations of the inputs $x$ and $x'$. Crucially, \citet{jacotNeuralTangentKernel2020} find that in the limit of infinite layer width, $\ntkk(x,x')$ becomes deterministic despite random weight initialization $\ntkk_0(x,x')=\ntkk(x,x')$ and remains constant throughout training, inducing analytically solvable training dynamics. As a result, the post-training NN function $f(x,\theta_\infty)$ can be characterized as a deterministic function of the random initialization $f(x,\theta_0)$ through
\begin{align}\label{ntkflin}
    f(x,\theta_\infty) = f(x,\theta_0) +\ntkk(x,\datax) \ntkk(\datax, \datax)^{-1}(\datay-f(\datax,\theta_0)) \,.
\end{align}
Here, we have overloaded notation to indicate the vectorization $\ntkk(x,\datax) \in \reals^{1 \times N_D}$, $\ntkk(\datax,\datax) \in \reals^{N_D \times N_D}$, and so forth. The matrix $\ntkk(\datax,\datax)$ is also known as the training Gram matrix, as we will refer to it. Further extending this framework, \citet{heBayesianDeepEnsembles2020a} demonstrate that by introducing suitable function priors on $f(x,\theta_0)$, akin to the well-known randomized prior functions by \citet{osbandDeepExplorationRandomized2019}, the post-training function is described by a Gaussian Process \citep[GP,][]{rasmussen2006gp}:
\begin{align} \label{ntkgp}
    f(\datax_t,\theta_\infty) \sim \mathcal{N} \big( \underbrace{\ntkk(\datax_t,\datax) \ntkk(\datax, \datax)^{-1} \datay}_{\mathbb E[f(\datax_t, \theta_\infty)]}, \,\,\,
    \underbrace{\ntkk(\datax_t, \datax_t) - \ntkk(\datax_t,\datax)\ntkk(\datax, \datax)^{-1}\ntkk(\datax, \datax_t)}_{\text{Cov}[f(\datax_t, \theta_\infty)]} \big) \,,
\end{align}
where $\datax_t$ is an arbitrary test data set. An outline of the derivation of Equations~\ref{ntkflin} and ~\ref{ntkgp} is provided in Appendix~\ref{sec:appendix1}. Consequently, the variance of an ensemble over infinite random initializations is given by 
\vspace{-2mm}
\begin{align}\label{ntkgpvar}
    \mathbb{V}[f(x,\theta_\infty)] = \ntkk(x,x)-\ntkk(x,\datax)\ntkk(\datax, \datax)^{-1}\ntkk(\datax, x) \,.
\end{align}
The above expression provides us with a theoretical footing for understanding the behavior and uncertainty estimates of deep ensembles. In the following sections we will describe our approach for estimating Eq.~\ref{ntkgpvar} not as the result of training several random models but deterministically with a single model.

\section{Contextual Similarity Distillation}
We now proceed to describe our approach, \textit{contextual similarity distillation} (CSD). The main objective of our method is to approximate the variance of an infinite deep ensemble, as described by Eq.~\ref{ntkgpvar}, directly with a single model. 

\subsection{Ensemble Variance Predictions for A Priori Queries} \label{sec:apriori}
We introduce the underlying idea of CSD in the simplified setting of \textit{a priori known} test points. Given a test query point $x_t$, it is our goal is to estimate the variance $\mathbb{V}[f(x_t,\theta_\infty)]$ of an ensemble of independently initialized NNs, trained on a dataset $\datax$. It is important to note that one could in principle obtain this variance via the NTK GP by solving Eq.~\ref{ntkgpvar}. This, however, requires inversion of the potentially very large Gram matrix $\ntkk(\datax, \datax)$, which becomes computationally prohibitive for most datasets and models of interest, including RL applications where sample sizes can go into the billions. 

\begin{figure}[t]
    \begin{center}
        \includegraphics[width=0.95 \columnwidth]{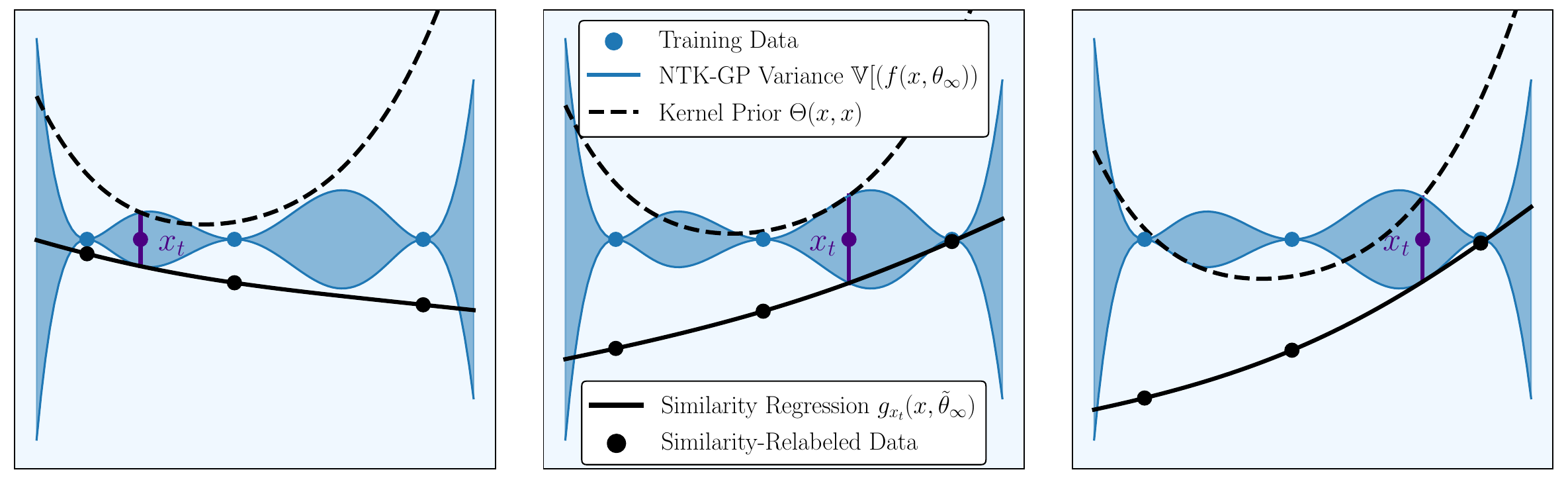}
        \vspace{0mm}
        \caption{ Illustration of regression tasks with query-dependent NTK similarities as labels. The difference between the kernel prior function $\ntkk(x,x)$ (dotted line) and the post-training regression function $g_{x_t}(x,\tiltheta_\infty)$ matches exactly ensemble variance in $x_t$. Plots from left to right depict the same principle, but for different query points $x_t$.}
        \label{fig:csdillustraion}
        \vspace{-0mm}
    \end{center}
\end{figure}

Instead of solving Eq.~\ref{ntkgpvar} directly, we leverage an alternative perspective that arises naturally from the learning dynamics of wide neural networks. Specifically, we begin with the simple observation that the variance of a wide ensemble at a test point $x_t$ can 
be computed efficiently as the solution to a regular supervised regression problem of a single model with a particular label function. For this, let $g(x,\tiltheta_t)$ be a NN of the same architecture as $f(x,\theta_t)$, thus inducing an equal NTK  $\ntkk_g(x,x') = \ntkk(x,x')$. Recall that the post-training NN function $g(x,\tiltheta_\infty)$ with squared loss on $\datay$ is given by 
\begin{align}
    g(x,\tiltheta_\infty) = g(x,\tiltheta_0) +\ntkk_g(x,\datax) \ntkk_g(\datax, \datax)^{-1}(\datay-g(\datax,\tiltheta_0)) \,.
\end{align}
It is straightforward to see that for small function initialization\footnote{For example, small function initialization can simply be obtained by redefining $\hat{f}(x,\theta_t) := f(x,\theta_t) - f(x,\theta_0)$.} $g(x,\tiltheta_0) \approx 0, \, \forall x$ the r.h.s. of this expression, when choosing the label function $\datay_{x_t}(\datax) = \ntkk(\datax, x_t)$, simplifies to 
\begin{align} \label{eq:gxt}
   g_{x_t}(x, \tiltheta_\infty) = \Theta(x,\datax)\Theta(\datax,\datax)^{-1}\Theta(\datax,x_t), 
\end{align}
where we used the subscript $x_t$ to indicate the function's dependence on the label function $\datay_{x_t}$. This identity now recovers exactly the problematic right term of Eq.~\ref{ntkgpvar} containing the Gram inversion $\ntkk(\datax, \datax)^{-1}$. Note that $g_{x_t}(x,\tiltheta_\infty)$ is obtained ``naturally'' as the result of gradient-based regression, without requiring explicit inversion of $\Theta(\datax,\datax)$ or training of a large ensemble at any point. The ensemble variance in a query point $x_t$ can then be obtained through the simple expression
\begin{align} \label{csdfixquery}
    \mathbb{V}[f(x_t, \theta_\infty)] = \ntkk(x_t,x_t) - g_{x_t}(x_t, \tiltheta_\infty),
\end{align}
which can be computed efficiently. Fig.~\ref{fig:csdillustraion} illustrates the above-described process of obtaining the expression~\ref{csdfixquery} geometrically. While simple, we believe this formulation provides a crucial insight: uncertainty estimation for a NN can be phrased as a singular prediction problem of kernel similarities. 

\subsection{Ensemble Variance Estimation for Arbitrary Query Points} \label{sec:anyquery}
In the above derivation, we outlined an efficient method for obtaining ensemble variances at a specific test query point $x_t$ \textit{known a priori}. An obvious limitation of this approach, however, is that the used labeling function $\datay_{x_t}(\datax) = \ntkk(\datax,x_t)$ and by extension the model $g_{x_t}(x, \tiltheta_\infty)$ is inherently dependent on the test point $x_t$ and not usable for arbitrary queries. 

To overcome this limitation, we now formulate a \textit{contextualized regression model} $g(x,c,\tiltheta_t)$, where $c$ serves as a context variable that determines the label function used during training of the function $g(x,c,\tiltheta_t)$. Specifically, instead of defining a label function that depends on a single fixed test query $x_t$, we construct a family of label functions parameterized by the context $c$, $\datay_c(\datax) = \ntkk(\datax, c)$. This means that for a set of context data $\datac = \{c_i \in \mathbb{R}^n | i\in\{1,...,N_C\}\}$, the model $g(x,c,\tiltheta_t)$ is optimized to solve a supervised regression problem associated with labels $\datay_c(\datax)$. 

Intuitively, this approach can be interpreted as an attempt to interpolate between multiple regression solutions that were trained on the same dataset $\datax$ but with different label functions $\datay_c(\datax)$. Geometrically, this corresponds to conjoining the functions $g_{x_t}$ in Fig.~\ref{fig:csdillustraion} along a new dimension $c$. So long as $g(x, c, \tiltheta_\infty)$ maintains the approximate dynamics of $g_c(x, \tiltheta_\infty)$, this model can be evaluated quickly for arbitrary test points by setting $c=x_t$ in
\begin{align} \label{csdapprox}
    g(x,c,\tiltheta_\infty) \approx \Theta(x,\datax)\Theta(\datax,\datax)^{-1}\Theta(\datax,c).
\end{align}
This generalization accordingly enables ensemble variance estimation across arbitrary points $x$ without requiring a separate regression solution for each individual query by computing
\begin{align} \label{eq:cntvar}
    \mathbb{V}[f(x,\theta_\infty)] \approx \ntkk(x, x) - g(x,x,\tiltheta_\infty).
\end{align} 
An intuitive interpretation of the function $g(x,x,\tiltheta_\infty)$ is that it captures an ensemble's confidence gained through observing the training data $\datax$, weighted by its similarity to $x$. The resulting variance of Eq. \ref{eq:cntvar} can then be understood as the difference between a prior uncertainty term $\ntkk(x,x)$ and the confidence term $g(x,x,\tiltheta_\infty)$. One should note at this point, that the evaluation of $g(x,c,\tiltheta_\infty)$ for contexts $c \notin \datac$ not used during training requires $g$ to generalize to novel $c$. Furthermore, the introduction of the context variable $c$ may influence the training dynamics of $g$, putting this approach into the realm of approximate algorithms. We have added a section to Appendix~\ref{sec:approx} that discusses and summarizes used approximations and their implications for practical settings.

\paragraph{Finetuning Variance Estimates with Context Data.} Before proceeding to describe our practical setup, we outline a property of contextualized similarity distillation that emerges through the above-described modeling choices. Our theoretical motivation highlights that \textit{exact} ensemble variances (in the NTK regime) can be obtained when the test point $x_t$ is known a priori. The implication of the subsequent formulation as a contextualized regression problem is that, when available, one can include unlabeled context data $\datac$ during training to obtain better uncertainty estimates in the domain of interest, as we will show later in the experimental section. This property also opens up the possibility of using unlabeled data augmentations to improve uncertainty estimation, an approach that has proven extraordinarily successful in the field of self-supervised and representation learning \citep{chen2020simple, guo2022byol, caron2021emerging} and not easily incorporated with standard approaches for uncertainty estimation \citep{lakshminarayananSimpleScalablePredictive2017, galDropoutBayesianApproximation2016a, burda2018exploration}.


\subsection{Contextualized Similarity Distillation with Deep Neural Networks} \label{sec:csdpractice}

\begin{figure}[t]
    \begin{center}
        \includegraphics[width=0.95 \columnwidth]{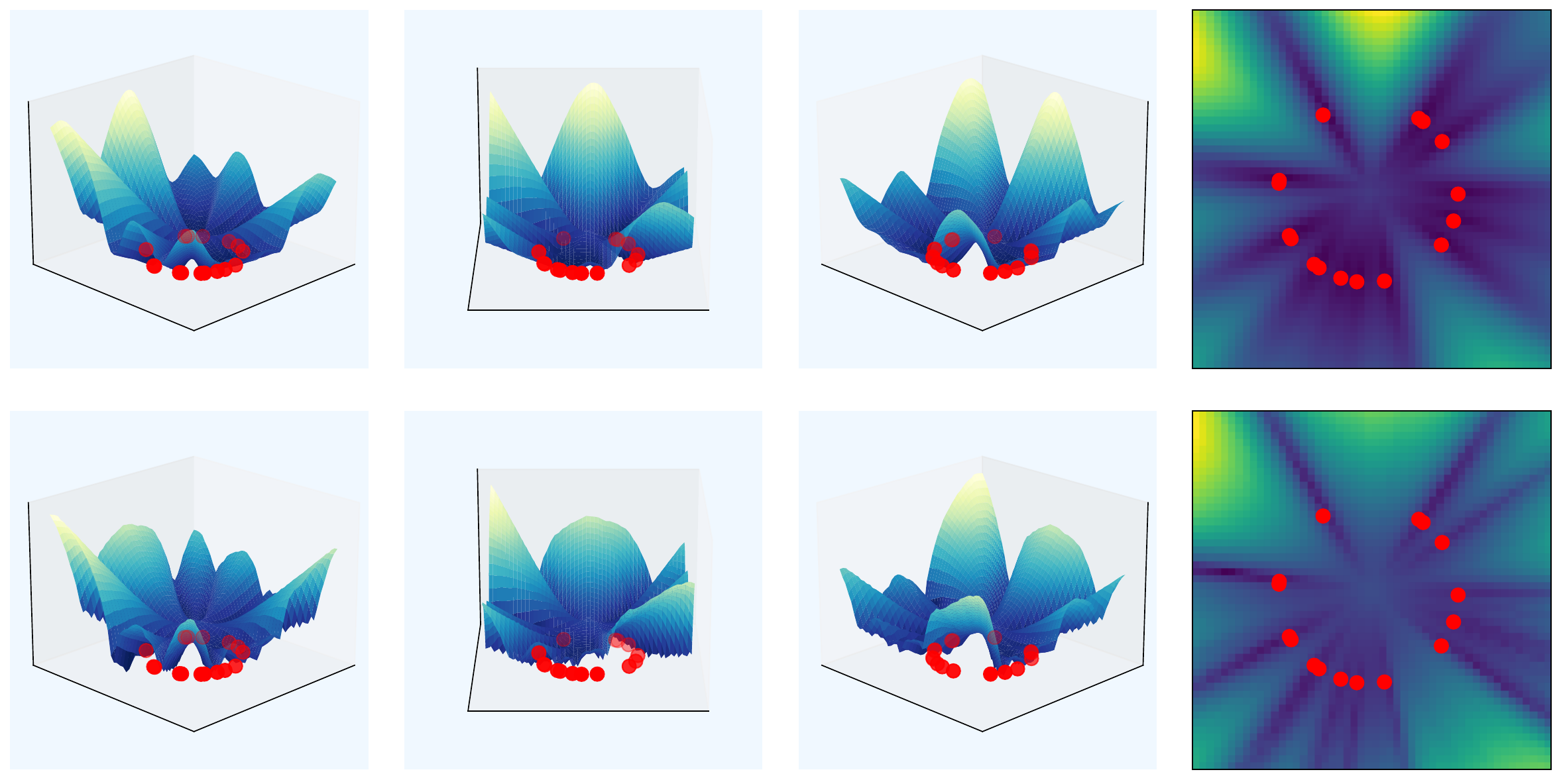}
        \vspace{0mm}
        \caption{\textit{Top Row}: Variance of an ensemble of 100 randomly initialized neural networks on a 2D toy regression task. Red dots are training points. \textit{Bottom Row}: Variance prediction by contextual similarity distillation (CSD) with a single model on the same regression task.}
        \label{fig:toyvariances}
        \vspace{-0mm}
    \end{center}
\end{figure}

Building on this theoretical basis, we proceed to describe a setting for contextualized similarity distillation with deep neural networks. This section outlines algorithmic design choices we found to be computationally efficient while maintaining the approach's theoretical motivation. 

First, we parameterize the contextualized regression model $g(x,c,\tiltheta_\infty)$ as an inner product between a feature vector $\phi(x, \tiltheta_f)$ and a context vector $\psi(c, \tiltheta_c)$ as
\begin{align}
    g(x,c,\tiltheta_\infty) = \phi(x, \tiltheta_f)^\top\psi(c, \tiltheta_c) \,.
\end{align}
Conceptually, this parametrization can be thought of as introducing a context-dependent final layer of weights, represented by $\psi(c, \tiltheta_c)$, to the regression model $g$. Computationally, this inner product parametrization bears the advantage that $g(\datax,\datac,\tiltheta_\infty) \in \reals^{N_D \times N_C}$ can be evaluated quickly without requiring explicit forward passes for each pairing $(x_i \in \datax,\, c_j \in \datac)$.

Second, we approximate the NTK prior $\ntkk(x,x')$ with partial gradients. Given that $\ntkk(x,x')$ is not involved in backward gradient computations, computing the full analytical or empirical prior kernel functions $\ntkk(x,x')$ is often not computationally prohibitive, but can pose a burden for models with large parameter spaces. We find that gradients with respect to only the last layer weights $\theta_{0}^L$ are sufficient in practice and further accelerate computation. Assuming, the last layer of $f$ is a dense layer such that $f(x,\theta_0) = \varphi(x, \theta_0^{1:L-1})^\top\theta_0^{L}$, we have
\begin{align}\label{partial-gradients}
    \ntkk^L(x,x') = \nabla_{\theta_0^{L}} f(x,\theta_0)^\top \nabla_{\theta_0^{L}} f(x',\theta_0) = \varphi(x, \theta_0^{1:L-1})^\top \varphi_f(x',\theta_0^{1:L-1}).
\end{align}
The resulting training pipeline for $g(x,c,\tiltheta_t)$ involves a simple supervised regression task with minimization of the squared loss
\begin{align} \label{csdloss}
    \mathcal{L} (\tiltheta_t) = \frac{1}{N}\sum_i^{N}\frac{1}{2} \big( g(x_i,c_i,\tiltheta_t) - \ntkk^L(x_i,c_i) \big)^2\,,
\end{align}
where $(x_i,c_i)$ are sampled randomly from $\datax$ and $\datac$. 

Lastly, we propose several choices for the context data $\datac$. We find that the arguably simplest choice, that is to reuse the training set $c_i \sim \datax$, works well in practice and is easily implemented. In addition, it is possible to apply data augmentations to the training samples $\datax$ when using as context data. For this, we employ the well-established set of augmentations from the contrastive learning literature \citep{chen2020simple}. We note here, that designing novel data augmentation techniques for the purpose of uncertainty quantification is a promising avenue (see for example works by \citet{wen2020combining} and \citet{wu2024posterior}). Unlike contrastive learning and many other self-supervised methods, our approach does not require data augmentations to preserve the nature of the original label and can in principle use any unlabeled data. Finally, when available, unlabeled data from the test distribution of interest can be used and often provides an additional improvement in uncertainty estimation, as we will show empirically.

\section{Empirical Evaluation}
Our empirical evaluation aims to provide us with a better understanding of contextual similarity distillation in practice. Given that our approach introduces approximations beyond the theoretical framework, we investigate whether CSD maintains its theoretically motivated properties in practice with high-dimensional problem and parameter spaces. Specifically, we aim to assess whether CSD provides a scalable alternative to deep ensembles and other established methods in uncertainty quantification, including Monte Carlo dropout \citep{galDropoutBayesianApproximation2016a}, a Bayesian NN based on Markov chain Monte Carlo sampling \citep[BNN - MCMC,][]{garriga2021exact}, a Laplace approximated Bayesian NN \citep[BNN - Laplace,][]{immer2021improving}, deep ensembles of sizes $3$ and $15$ \citep[ENS,][]{lakshminarayananSimpleScalablePredictive2017} and random network distillation \citep[RND,][]{burda2018exploration}. Furthermore, we analyze how algorithmic design choices, such as the choice of context data, influence uncertainty estimates. Lastly, we seek to evaluate our approach's efficacy as an exploration signal for deep reinforcement learning agents on sparse-reward visual exploration tasks from the VizDoom~\citep{Kempka2016ViZDoom} suite. 

\subsection{Distribution Shift Detection}
Following prior work \citep{van2020uncertainty, immer2021improving, rudner2022tractable}, we evaluate uncertainty estimates in an image classification setting under distribution shift, where a model trained on an in-distribution dataset is evaluated on inputs from a shifted distribution. 

\begin{table}
  \caption{Distribution Shift Detection. Test accuracy and average OOD detection metrics across MNIST, FashionMNIST, KMNIST, NotMNIST. OOD metrics are evaluated for each ID dataset against the remaining OOD datasets and a perturbed version of the ID dataset.}
  \label{tab:ooddetection}
  \centering
    \begin{tabular}{lllll}
        \toprule
        \textbf{Method} & \textbf{Acc.} & \textbf{AUROC} & \textbf{AUPR-IN} & \textbf{AUPR-OUT} \\
        \midrule
        MCD & $94.39\pm 0.10$ & $85.67\pm 0.21$ & $81.73\pm 0.34$ & $86.44\pm 0.20$ \\
        BNN-MCMC & $87.70\pm 0.38$ & $83.17\pm 0.60$ & $82.65\pm 0.66$ & $82.28\pm 0.71$ \\
        BNN-Laplace & $90.86\pm 0.62$ & $81.38\pm 0.73$ & $79.43\pm 0.84$ & $81.84\pm 0.66$ \\
        RND & $96.18\pm 0.05$ & $\bm{94.40}\pm 0.41$ & $94.17\pm 0.63$ & $\bm{94.01}\pm 0.31$ \\
        ENS(3) & $96.91\pm 0.04$ & $92.30\pm 0.09$ & $92.83\pm 0.10$ & $91.37\pm 0.11$ \\
        ENS(15) & $\bm{97.18}\pm 0.03$ & $94.00\pm 0.07$ & $\bm{94.70}\pm 0.07$ & $92.99\pm 0.06$ \\
        \midrule
        CSD & $96.29\pm 0.07$ & $\bm{96.63}\pm 0.35$ & $\bm{96.94}\pm 0.39$ & $\bm{96.19}\pm 0.32$ \\
        CSD-Aug. & $96.28\pm 0.06$ & $\bm{98.22}\pm 0.14$ & $\bm{98.51}\pm 0.13$ & $\bm{97.80}\pm 0.17$ \\
        CSD-OOD. & $96.30\pm 0.06$ & $\bm{98.57}\pm 0.14$ & $\bm{98.86}\pm 0.12$ & $\bm{98.19}\pm 0.15$ \\
        \bottomrule
    \end{tabular}
\end{table}

In particular, we train models on one of the FashionMNIST, MNIST, KMNIST, NotMNIST datasets and evaluate uncertainty estimates on the other, shifted datasets and a perturbed version of the in-distribution dataset. Well-calibrated epistemic uncertainty estimates will correlate with dataset shift, such that out-of-distribution samples are likely to be rated more uncertain than in-distribution samples. To compare methods quantitatively, we use the threshold-independent area under the receiver operating characteristic curve (AUROC) metric, as well as the area under the precision-recall curve for in-distribution (AUPR-IN) and out-of-distribution (AUPR-OUT) samples. The AUROC metric can be interpreted as the likelihood of an OOD sample receiving higher uncertainty than an ID sample, while AUPR-IN and AUPR-OUT provide additional sensitivity to dataset size and the choice of the positive class. For these metrics, Table~\ref{tab:ooddetection} reports the average and standard deviation over $10$ seeds, averaged over all permutations of ID and OOD datasets, along with average test accuracy. Full detailed results are provided in the supplementary material. 

To analyze the role of the used context data, we evaluate three versions of CSD: a baseline that only uses training data (CSD), a variant incorporating data augmentations to training samples (CSD-Aug.), and a model using context data from the evaluation distribution (CSD-OOD). Even in the basic version, CSD demonstrates highly effective distribution shift detection, surpassing baseline methods on a variety of datasets while requiring only a single model. Our results furthermore suggest that incorporating data augmentations and target-distribution context data indeed significantly improves performance.

\subsection{Exploration in VizDoom}
We now evaluate CSD in a reinforcement learning task with high-dimensional observation spaces and sparse rewards. For this, we consider visual navigation tasks in the VizDOOM environment, where agents explore a 3D maze-like environment with ego-perspective image observations. The agent is tasked with reaching a goal while receiving no reward signal except upon successful completion. We consider three variations of the task, where agents are initialized at increasing distances from the goal, defining progressively harder exploration tasks. 
\begin{figure}[t]
    \begin{center}
        \hspace*{-2mm}\import{images}{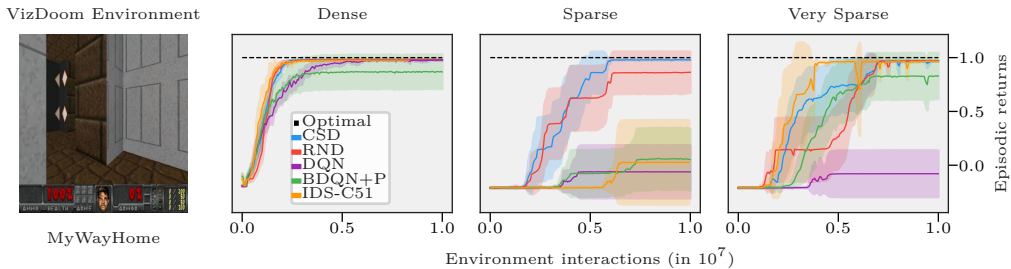}
        \caption{(Left): Visual observation in the VizDoom environment \citep{Kempka2016ViZDoom}. (From Second Left to Right): Mean learning curves in variations of the \textit{MyWayHome} VizDoom environment. Shaded regions are $90\%$ Student’s t confidence intervals from 10 seeds.}
        \label{fig:vizdoom}
        \vspace{-5mm}
    \end{center}
\end{figure}

We use Rainbow \citep{hessel2018rainbow} as a base algorithm and include uncertainty estimates by CSD as an intrinsic reward (full details provided in Appendix~\ref{sup:exp}). For a comparative evaluation, we compare the performance of CSD-based exploration with several baseline algorithms, including deep Q networks \citep[DQN,][]{mnihPlayingAtariDeep2013}, random network distillation \citep[RND,][]{burda2018exploration}, bootstrapped Q-networks \citep[BDQN+P,][]{osbandDeepExplorationRandomized2019}, and information-directed sampling \citep[IDS,][]{nikolovInformationDirectedExplorationDeep2019}. Fig.~\ref{fig:vizdoom} shows mean learning curves across 10 random seeds. Interestingly, the sparse version of the environment appears to be the hardest, a circumstance we believe is due to the spawning point lying in a sidearm of the maze map. Of the tested methods, only CSD was able to find the goal across all seeds and environments, with RND performing most competitively.

\section{Related Work}
Our work builds on the extensive body of literature in the field of uncertainty quantification in deep learning and reinforcement learning. Ensemble learning~\citep{dietterich2000ensemble} has emerged as on the most effective and reliable approaches to uncertainty estimation  \citep{lakshminarayananSimpleScalablePredictive2017} and has been widely adopted in the deep reinforcement learning literature. In particular, ensembles can be used for efficient exploration by sampling random models \citep{osbandDeepExplorationBootstrapped2016b, qin2022analysis, osbandWhyPosteriorSampling2017}, by constructing upper confidence bounds for exploration bonuses \citep{chen2017ucb, odonoghueUncertaintyBellmanEquation2018} or by estimating information gain \citep{nikolovInformationDirectedExplorationDeep2019}. Several works moreover rely on deep ensembles to reduce overestimation and improve learning stability \citep{fujimotoAddressingFunctionApproximation2018b, haarnoja2018soft, chenRandomizedEnsembledDouble2021}, extending to the challenging offline setting \citep{an2021uncertainty, agarwal2020optimistic, smit2021pebl}. 

A number of previous works have focused on reducing ensemble size, notably by disaligning the Jacobian of networks \citep{an2021uncertainty}, adding repulsive loss terms \citep{sheikh2022maximizing}, or through architectural diversification \citep{osbandDeepExplorationRandomized2019, zanger2024diverse}. Notably, various works aim to quantify epistemic uncertainty with a single model \citep{pathakCuriosityDrivenExplorationSelfSupervised2017, burda2018exploration, filos2021model, guo2022byol, lahlou2021deup}, often by measuring prediction errors. To the best of our knowledge, few single-model methods in the field offer an interpretation as ensemble or posterior uncertainty. 

In a broader sense, ensembles have been studied extensively from a Bayesian perspective \citep{hoffmann2021deep, dangelo2021repulsive}. In particular, some of our work relies on the NTK GP characterization of deep ensembles by \citet{heBayesianDeepEnsembles2020a}, who, in turn, rely on seminal work by seminal work on the NTK by \citet{jacotNeuralTangentKernel2020} and \citet{leeWideNeuralNetworks2020}. Subsequent analysis has used the NTK to disentangle ensemble variance \citep{kobayashi2022disentangling}. Recent works \citet{wilson2025uncertainty} rely on NTK theory to derive a sampling-based uncertainty estimator, while \citet{calvo2024epistemic} construct uncertainty estimates using several regression models. In contrast to the latter, our method uses a contextualized regression model that allows for single-model uncertainty estimates in a deep learning setting.

\section{Conclusion}
This work introduced \textit{contextual similarity distillation} (CSD), a novel single-model approach for uncertainty quantification that estimates the predictive variance of an ensemble with a single model and forward pass. By reframing ensemble variance estimation as a structured regression problem, CSD enables efficient uncertainty estimation without requiring the training of multiple models, stochastic forward passes, or explicit kernel matrix inversion. Instead, phrasing predictive variance estimation as a contextualized regression problem is amenable to standard training pipelines with deep NNs and gradient descent. 

We implemented CSD in a deep learning setting and performed a comparative evaluation on a variety of distribution shift detection and reinforcement learning tasks. Empirically, we found that CSD provides uncertainty estimates competitive and sometimes superior to deep ensembles and other alternatives on all tasks. This makes CSD an attractive option for guiding exploration in RL, as our experiments on high-dimensional exploration tasks confirmed. Our results furthermore confirmed that our approach can leverage unlabeled target domain data and data augmentations to further refine uncertainty estimates. We believe our work opens up several avenues for future research, including applications in model-based and offline RL, or the use of more refined data augmentation techniques.

Our findings, we believe, position CSD as a scalable alternative to deep ensembles, offering a principled and computationally efficient method for uncertainty quantification in deep learning. 

\appendix
\section{Linearized Neural Network Learning Dynamics}
\label{sec:appendix1}
For completeness, we briefly outline a sketch for how the GP interpretation of wide neural networks governed by NTK dynamics described in Expression~\ref{ntkgp} can be obtained. This section largely follows the seminal works by \citet{jacotNeuralTangentKernel2020}, \citet{leeWideNeuralNetworks2020} and \citet{heBayesianDeepEnsembles2020a}, to whom we refer readers interested in further details. 

We begin by constructing a first-order Taylor expansion of the neural network function $f(x, \theta_0)$ around its initialization parameters $\theta_0$: 
\begin{align}
   \flin(x, \theta_t) = f(x, \theta_0) + \nabla_{\theta} f(x, \theta_0)^\top (\theta_t - \theta_0). 
\end{align}
When trained on $\datax$ and $\datay$ with the squared error loss $\mathcal{L} = \frac{1}{2} \| \flin(\mathcal{X};\theta_t)-\mathcal{Y}\|^2$, gradient flow with a learning rate $\alpha$ induces an evolution of $\theta_t$ according to 
\begin{align}
    \dv{t} \theta_t &= -\alpha \nabla_\theta \mathcal{L} = -\alpha  \nabla_{\theta}\flin(\datax, \theta_t) \nabla_{\flin(\datax, \theta_t)} \mathcal{L} \,.
\end{align}
In function space, this evolution translates to the expression
\begin{align} \label{flinode}
     \dv{t} \flin(x;\theta_t) &= \nabla_\theta \flin(x, \theta_t)^\top \dv{t} \theta_t = -\alpha \ntkk_0(x,\datax) (\flin(\mathcal{X};\theta_t)-\mathcal{Y}) \,,
\end{align}
where $\ntkk_0(x,x') = \nabla_{\theta} f(x, \theta_0)^\top \nabla_{\theta} f(x', \theta_0)$ is the (empirical) tangent kernel of $\flin(x,\theta_t)$. Since this linearization has constant gradients $\nabla_{\theta} f(x, \theta_0)$, the resulting differential equation is linear and solvable. For the substitution $v_t = (\flin(\mathcal{X};\theta_t)-\mathcal{Y})$, we obtain the training error dynamics $\dv{t} v_t = -\alpha \ntkk_0(\datax, \datax)v_t$
to which an exponential ansatz yields the solution
\begin{align} \label{flintraindyn}
    \flin(\mathcal{X};\theta_t) -\mathcal{Y} &= e^{-\alpha t \ntkk_0(\datax, \datax)} (f(\mathcal{X};\theta_0)  -\mathcal{Y}) \,,
\end{align}
where the matrix exponential $e^{-\alpha t \ntkk_0 (\datax, \datax)}$ was used. Plugging Eq.~\ref{flintraindyn} back into Eq.~\ref{flinode}, one arrives at the identity 
\begin{align}
     \dv{t} \flin(x;\theta_t) &= -\alpha \ntkk_0(x,\datax) e^{-\alpha t \ntkk_0(\datax, \datax)} (f(\mathcal{X};\theta_0)-\mathcal{Y}) \,.
\end{align}
This differential expression is explicit in its terms such that we can obtain a solution by integration through
\begin{align}
    \flin(x;\theta_t) &= f(x,\theta_0) + \int_0^t \dv{t'} \flin(x, \theta_{t'}) \dd t' \\
    &= f(x,\theta_0) + \ntkk_0(x, \datax) \ntkk_0(\datax,\datax)^{-1}(e^{-\alpha t \ntkk(\datax,\datax)}-I)(f(\datax, \theta_0) - \datay) \, ,
\end{align}
which recovers Eq.~\ref{ntkflin} for $t \xrightarrow{} \infty$. A central result by \citet{jacotNeuralTangentKernel2020} and extended in the linearized setting by \citet{leeWideNeuralNetworks2020} is that, as layer widths of the neural network go to infinity, the NTK $\ntkk_0(x,x')$ becomes deterministic and constant and the linear approximation $\flin(x;\theta_t)$ becomes exact w.r.t. the original function $\lim_{\text{width}\xrightarrow{} \infty}\flin(x;\theta_t)=f(x,\theta_t)$. 

Rewriting the (infinite width) post-training test and training functions as an affine transformation of the initialization yields
\begin{align} \label{blockeq}
    \begin{pmatrix}
        f(\datax_t,\theta_\infty) \\
        f(\datax,\theta_\infty)
    \end{pmatrix} 
    &=
    \begin{pmatrix}
        1 & -\ntkk(\datax_t, \datax)\ntkk(\datax,\datax)^{-1} \\
        0 & 0
    \end{pmatrix} 
    \begin{pmatrix}
        f(\datax_t,\theta_0) \\
        f(\datax,\theta_0)
    \end{pmatrix}
    +
    \begin{pmatrix}
        \ntkk(\datax_t, \datax)\ntkk(\datax, \datax)^{-1}\datay \\
        \datay
    \end{pmatrix} \,.
\end{align}
For the earlier described parametrization of $f$, the set of initial predictions is known to follow a multivariate Gaussian distribution \citep{lee2017deep} described by the neural network Gaussian process (NNGP) $f(\datax, \theta_0) \sim \mathcal{N}(0, \nngpk(\datax, \datax ))$ (and analogously for $\datax_t$), where
\begin{align}\label{nngp}
   \nngpk(\datax_t, \datax_t) = \mathbb{E}_{\theta_0} \left[ f(\datax_t, \theta_0) f(\datax_t, \theta_0)^\top \right] \,. 
\end{align}
Affine transformations of multivariate Gaussian random variables $X \sim \mathcal{N}(\mu_X, \Sigma_X)$ with $Y = a + BX$ are, in turn, multivariate Gaussian random variables with distribution $Y \sim \mathcal{N}(a+B\mu_X, \,\, B\Sigma_XB^\top)$. We here omit explicit  derivations and rearrangements for brevity. As a consequence, Eq.~\ref{blockeq} with initialization covariance from Eq.~\ref{nngp} is also described by a multivariate Gaussian with mean and covariance given by
\begin{align} \label{flincov} \nonumber 
    \mathbb{E}_{\theta_0}[f(\datax_t, \theta_\infty)] &= \ntkk(\datax_t, \datax) \ntkk(\datax_, \datax)^{-1}\datay \,, \\
    \text{Cov}(f(\datax_t, \theta_\infty)) &= \nngpk(\datax_t, \datax_t)-\ntkk(\datax_t,\datax)\ntkk(\datax,\datax)^{-1}\nngpk(\datax,\datax)\ntkk(\datax,\datax)^{-1}\ntkk(\datax,\datax_t) \\  \nonumber
    & \quad - (\ntkk(\datax_t,\datax)\ntkk(\datax,\datax)^{-1}\nngpk(\datax,\datax_t) + \text{h.c.}) \,,
\end{align}
where h.c. refers to the Hermitian conjugate of the preceding term. \citet{heBayesianDeepEnsembles2020a} then introduce constant ``correction'' terms to the function initialization described in Eq.~\ref{nngp}, in particular such that $\nngpk(x,x') = \ntkk(x,x')$. This simplifies Expression~\ref{flincov} significantly and now permits a Gaussian process interpretation with the final expression given by Eq.~\ref{ntkgp}.

\section{Discussion on Approximations}
\label{sec:approx}
As our method relies on several approximations, we include a discussion that aims to provide an overview of the approximate nature of our method and in which settings it is exact or where deviations may be more likely. 

The first central approximation we make is to model neural networks with dynamics governed by a deterministic and constant NTK. \citet{jacotNeuralTangentKernel2020} show that this is the case for fully connected NNs with NTK parametrization trained on a squared loss. The implied dynamics are solved assuming gradient flow, that is with infinitesimal step sizes and full-batch gradients. \citet{jacotNeuralTangentKernel2020} and \citet{leeWideNeuralNetworks2020}
moreover show that convergence and final generalization behavior is empirically well-described by wide but finite architectures including fully connected NNs, convolutional NNs and residual architectures, trained with stochastic gradient descent. The function initialization scheme proposed by \citet{heBayesianDeepEnsembles2020a} allows for a Gaussian process interpretation of NNs from random initialization and largely relies on the same assumptions as the above-described works. 

Our theoretical motivation, outlined in Sections~\ref{sec:apriori} and \ref{sec:anyquery}, relies on the GP description of deep ensembles and the implied assumptions. Given this setting, that is assuming NTK parametrization with infinite widths, function initialization according to \citet{heBayesianDeepEnsembles2020a}, and gradient flow with squared loss, the derivation for single-query ensemble variances in Section~\ref{sec:apriori} is exact. In our contextualized model described in Section~\ref{sec:anyquery}, we introduce an additional approximation through the introduction of an explicit context variable $c$, which may interfere with the training dynamics of $g(x,c,\tiltheta)$. Let training tuples be $x^c = (x,c)$ and $\datax^c = \{x_1^c, x_2^c,...,x_{N_{T}}^c\}$ and let the NTK of $g$ be $\ntkk_g((x,c),(x',c')) = \nabla_{\tiltheta} g(x,c,\tiltheta_0)^\top \nabla_{\tiltheta} g(x',c',\tiltheta_0)$. The analogous regression solution to the function $g(x,c,\tiltheta)$ by minimizing the loss in Eq.~\ref{csdloss} becomes 
\begin{align}
    g(x,c,\tiltheta_\infty) = \Theta_g(x^c,\datax^c)\Theta_g(\datax^c,\datax^c)^{-1}\Theta(\datax^c).
\end{align}
A natural setting in which these training dynamics recover Eq.~\ref{csdapprox} is when gradients are independent between context, that is $\Theta_g((x,c),(x,c'))=0$ if $c \neq c'$ and maintain the gradient structure of $\ntkk(x,x')$ with $\Theta_g((x,c),(x',c))=\ntkk(x,x'),\, \forall c \in \datac$. As this setting would hardly permit meaningful interpolations and extrapolations between different contexts $c$, one engages in a trade off between generalization capability towards general contexts $c$ and interference in the training dynamics. 

In our practical setting, we furthermore approximate the NTK prior function with partial gradients as outlined in Eq.~\ref{partial-gradients} of Section~\ref{sec:csdpractice}. The influence of this approximation choice generally depends on architecture, but we found it to perform well in our experiments using deep convolutional and residual architectures. Lastly, the RL exploration setting involves data streams rather than fixed datasets $\datax$, further deviating from the earlier delineated dynamics. Understanding the influence of this non stationarity on training dynamics is an open problem, and we believe countermeasures like periodic resets \citep{d2022sample} are a promising avenue for future research. 




\bibliography{main}
\bibliographystyle{rlj}

\clearpage
\clearpage


\beginSupplementaryMaterials

\section{Experimental Details} \label{sup:exp}
In the following, we outline details on our experimental setup. This includes hyperparameter settings, hyperparameter search procedures, algorithmic and experimental details, and dataprocessing details.

\subsection{Hyperparameter Settings}
In order to facilitate comparable results, our experiments are conducted using a central codebase and  follow similar modeling choices such as architectures, optimizer, etc. where sensible. All experiments use a resnet-based model \citep{he2016deep} following the IMPALA architecture by \citet{espeholt2018impala}. We optimized essential and algorithm-specific hyperparameters through a search on a selected subset of experiments.

\paragraph{Distribution shift detection.} In the supervised distribution shift detection, we use the IMPALA architecture with $2$ residual blocks and channels widths $32$ and $64$. Hyperparameters were searched on the FashionMNIST dataset as the in-distribution set and the remaining datasets as out-of-distribution sets. Each dataset is normalized to zero-mean and standard deviation $1$ using the training set statistics. For the main classifier we apply random horizontal flips (p=0.5), random vertical flips (p=0.5) and random sized crops (zoom range between 1.0 and 1.3) to training data in all experiments. Learning rate and algorithm-specific hyperparameters were optimized independently, meaning we first performed a search for learning rates, which we used in the (if applicable) subsequent algorithm-specific parameter searches. Table~\ref{hpsearchspaceood} contains lists of all searched parameters, with parenthesis indicating algorithm-specific parameters and italics indicating the parameter used during the learning rate search. The final hyperparameters were chosen based on the average AUROC metric and are reported in Table~\ref{hpsettingsood}.

\begin{table}[p]
  \caption{Searched hyperparameters for distribution shift experiments.}
  \centering
  \begin{tabular}{ll}
    \toprule
    Hyperparameter     & Values \\
    \midrule
    Learning rate (All) & $[10^{-4}, \,3 \cdot 10^{-4}, \, 10^{-3},\,3 \cdot 10^{-3},\, 10^{-2},\, 3 \cdot 10^{-2},\, 10^{-1}]$ \\
    Dropout probability (MCD)     & $[0.05,\, 0.1,\, \mathit{0.15},\, 0.25,\, 0.5]$ \\
    RND Learning rate (RND) & $[10^{-4}, \,3 \cdot 10^{-4}, \, \mathit{10^{-3}},\,3 \cdot 10^{-3},\, 10^{-2},\, 3 \cdot 10^{-2},\, 10^{-1}]$ \\
    CSD Learning rate (CSD) & $[10^{-4}, \,3 \cdot 10^{-4}, \, \mathit{10^{-3}},\,3 \cdot 10^{-3},\, 10^{-2},\, 3 \cdot 10^{-2},\, 10^{-1}]$ \\
    \bottomrule
  \end{tabular}
  \label{hpsearchspaceood}
\end{table}
\begin{table}[p]
  \caption{Hyperparameter settings for distribution shift experiments.}
  \label{hpsettingsood}
  \centering
  \begin{tabular}{lllllll}
    \toprule
    Hyperparameter & MCMC & Laplace & MCD & ENS & RND &  CSD\\
    \midrule
    \multicolumn{7}{c}{Main Classifier Network}\\
    \midrule
    Learning rate & $10^{-3}$ & $10^{-3}$ & $3\cdot10^{-4}$ & $10^{-3}$ & $10^{-3}$ & $10^{-3}$\\
    MLP hidden layers & \multicolumn{6}{c}{$2$}\\
    MLP layer width & \multicolumn{6}{c}{$256$} \\
    Channel Widths & \multicolumn{6}{c}{$32, \,64$}\\
    \midrule
    \multicolumn{7}{c}{RND/CSD Network}\\
    \midrule
    Learning rate & \multicolumn{4}{c}{n/a} & $3 \cdot 10^{-3}$ & $10^{-2}$\\
    MLP hidden layers & \multicolumn{4}{c}{n/a} & $2$ & $2$\\
    MLP layer width & \multicolumn{4}{c}{n/a} & $256$ & $256$\\
    Channel Widths & \multicolumn{4}{c}{n/a} & $16$ & $32$\\
    Target hidden layers & \multicolumn{4}{c}{n/a} & $1$ & $1$\\
    Output dimensions & \multicolumn{4}{c}{n/a} & $256$ & $256$\\
    \midrule
    Ensemble size & n/a & n/a & n/a & $3, \,15$ & n/a & n/a\\
    Dropout rate & n/a & n/a & $0.1$ & \multicolumn{3}{c}{n/a} \\
    Prior Precision & n/a & 100 & n/a & \multicolumn{3}{c}{n/a} \\
    Posterior Temperature & 1.0 & 1.0 & n/a & \multicolumn{3}{c}{n/a} \\
    Posterior Samples & 30 & 30 & 100 & \multicolumn{3}{c}{n/a} \\
    Epochs per sample & 2 & n/a & n/a & \multicolumn{3}{c}{n/a} \\
    Burn-In Epochs & 10 & n/a & n/a & \multicolumn{3}{c}{n/a} \\
    Adam epsilon & n/a & $10^{-5}$ & $10^{-5}$ & \multicolumn{3}{c}{$10^{-5}$} \\
    Learning rate anneal & \multicolumn{5}{c}{Linear} \\
    Batch size & \multicolumn{5}{c}{$256$}\\ 
    Initialization & \multicolumn{5}{c}{Orthogonal \citep{saxe2013exact}} \\
    \bottomrule
  \end{tabular}
\end{table}

\paragraph{VizDoom.} In the RL experiments, we conducted a full grid search on the \textit{MyWayHomeSparse} variation of the environment and chose parameters based on performance after $5 \cdot 10^6$ steps. Our basic network architecture is based on the rainbow network proposed by \citet{schmidt2021fast} who in turn base their architecture on IMPALA \citep{espeholt2018impala}. We use 3 residual blocks with channel widths according to Table~\ref{vizdoompreproc}. Detailed final hyperparameter settings are given in Table~\ref{hpsettingsvizdoom}. All agents furthermore use a data preprocessing pipeline as outlined in Table~\ref{vizdoompreproc}.

\begin{table}[p]
  \caption{Searched hyperparameters for VizDoom}
  \centering
  \begin{tabular}{ll}
    \toprule
    Hyperparameter     & Values \\
    \midrule
    Learning rate     & $[1.25 \cdot 10^{-4}, 2.5 \cdot 10^{-4}, 3.75 \cdot 10^{-4},$ \\ & $ 5 \cdot 10^{-4}, 6.25 \cdot 10^{-4}, 7.5 \cdot 10^{-4}]$ \\
    Prior function scale (BDQN+P, IDS) & $[1.0, 3.0, 5.0]$ \\
    Initial bonus $\beta$ (RND, CSD) & [0.05, 0.1, 0.5, 1.0, 5.0, 10.0] \\
    RND Learning rate (RND) & $[1.25 \cdot 10^{-4}, 2.5 \cdot 10^{-4}, 3.75 \cdot 10^{-4},$ \\ & $ 5 \cdot 10^{-4}, 6.25 \cdot 10^{-4}, 7.5 \cdot 10^{-4}]$ \\
    CSD Learning rate (CSD) & $[1.25 \cdot 10^{-4}, 2.5 \cdot 10^{-4}, 3.75 \cdot 10^{-4},$ \\ & $ 5 \cdot 10^{-4}, 6.25 \cdot 10^{-4}, 7.5 \cdot 10^{-4}]$ \\
    \bottomrule
  \end{tabular}
  \label{hpsearchspacevizdoom}
\end{table}

\begin{table}[p]
  \caption{Hyperparameter settings for VizDoom experiments.}
  \label{hpsettingsvizdoom}
  \centering
  \begin{tabular}{llllll}
    \toprule
    Hyperparameter & DQN & BDQN+P & RND & IDS & CSD \\
    \midrule
    Adam Learning rate & $2.5 \cdot 10^{-4}$ & $2.5 \cdot 10^{-4}$ & $6.25 \cdot 10^{-4}$ & $2.5 \cdot 10^{-4}$ & $6.25 \cdot 10^{-4}$\\
    Prior function scale & n/a & 1.0 & n/a & 1.0 & n/a \\
    Heads $K$ & $1$ & $1$ & $101$ & $1$ / $101$ & $101/101$ \\
    Ensemble size & n/a & $10$ & n/a & $10/1$ & n/a \\
    Initial bonus $\beta_{\text{init}}$ & n/a & n/a & $1.0$ & $0.1$ & $0.1$ \\
    Final bonus $\beta_{\text{final}}$ & n/a & n/a & $0.01$ & $0.01$ & $0.01$ \\
    Bonus decay frames & n/a & n/a & $3.3 \cdot 10^6$ & $3.3 \cdot 10^6$ & $3.3 \cdot 10^6$ \\
    Loss function & Huber & Huber & C51 & Huber/C51 & C51 \\
    Channel Widths & \multicolumn{5}{c}{$32, \, 32, \, 64$} \\
    MLP hidden layers & \multicolumn{5}{c}{$1$} \\
    MLP layer width & \multicolumn{5}{c}{$256$} \\
    \midrule
    \multicolumn{6}{c}{RND / CSD Network Parameters} \\
    \midrule
    Adam Learning rate & n/a & n/a & $2.5 \cdot 10^{-4}$ & n/a & $2.5 \cdot 10^{-4}$\\
    Channel Widths & n/a & n/a & $16,\,16,\,32$ & n/a & $16,\,16,\,32$ \\
    MLP hidden layers & n/a & n/a & $1$ & n/a & $1$\\
    MLP layer width & n/a & n/a & $256$ & n/a & $256$\\
    Target hidden layers & n/a & n/a & $1$ & n/a & $1$ \\
    Output dimensions & n/a & n/a & $256$ & n/a & $256$ \\
    \midrule
    Initial $\epsilon$ in $\epsilon$-greedy & \multicolumn{5}{c}{$1.0$}\\
    Final $\epsilon$ in $\epsilon$-greedy & \multicolumn{5}{c}{$0.01$}\\    
    $\epsilon$ decay frames & \multicolumn{5}{c}{$500,000$}\\
    Training starts & \multicolumn{5}{c}{$100,000$}\\
    Discount & \multicolumn{5}{c}{$0.997$}\\
    Buffer size & \multicolumn{5}{c}{$1,000,000$} \\
    Batch size & \multicolumn{5}{c}{$256$} \\
    Parallel Envs & \multicolumn{5}{c}{$16$}\\
    \midrule
    Adam epsilon & \multicolumn{5}{c}{$0.005 / \text{batch size}$} \\
    Initialization & \multicolumn{5}{c}{He uniform\ \citep{he2015delving}} \\
    Gradient clip norm & \multicolumn{5}{c}{10} \\
    Regularization & \multicolumn{5}{c}{spectral normalization \citep{gogianu2021spectral}} \\
    Double DQN & \multicolumn{5}{c}{Yes \citep{vanhasseltDeepReinforcementLearning2015}} \\
    Update frequency & \multicolumn{5}{c}{$1$} \\
    Target lambda & \multicolumn{5}{c}{$1.0$} \\
    Target frequency & \multicolumn{5}{c}{$8000$} \\
    PER $\beta_0$ & \multicolumn{5}{c}{$0.45$ \citep{schaul2015prioritized}} \\
    n-step returns & \multicolumn{5}{c}{$10$} \\
    \bottomrule
  \end{tabular}
\end{table}

\begin{table}[p]
  \caption{VizDoom Preprocessing}
  \centering
  \begin{tabular}{ll}
    \toprule
    Parameter & Value\\
    \midrule
    Grayscale    & Yes \\
    Frame-skipping & No \\
    Frame-stacking & 6 \\
    Resolution & $42\times42$ \\
    Max. Episode Length & 2100 \\
    \bottomrule
  \end{tabular}
  \label{vizdoompreproc}
  \vspace{-2mm}
\end{table}

\subsection{Implementation Details} \label{sup:impdets}
In this section, we briefly outline implementation details concerning CSD and the tested baselines. 
\paragraph{Data augmentations} For both the distribution shift detection experiments (CSD-Aug.) and the VizDoom experiments, we add data augmentation to obtain additional context variables in CSD. In both experiments, we apply augmentations with a probability of $p=0.25$ and specific augmentations are listed in Table~\ref{tab:dataaug}. 
\paragraph{Data and context sampling.} To compute the loss~\ref{csdloss}, we sample minibatches $\datax_{mb}$ from a buffer or data set. Context minibatches $\datac_{mb}$ either simply reuse $\datax_{mb}$, are generated by applying data augmentations as outlines above, or by sampling from a context data set. We compute inner products over all pairings of the two batches with $\phi(\datax_{mb}, \tiltheta_f)^\top\psi(\datac_{mb}, \tiltheta_c) \in \reals^{N_{mb} \times N_{mb}}$ and compute loss~\ref{csdloss} elementwise. Finally, we sum the average diagonal loss and the average off-diagonal loss. 
\paragraph{Normalization.} During training, we normalize prior features $\bar{\varphi}(x, \theta_0^{1:L-1}) = \varphi(x, \theta_0^{1:L-1}) / \|\varphi(x, \theta_0^{1:L-1}) \|_2$, feature vectors $\bar{\phi}(x, \tiltheta_f) = \phi(x, \tiltheta_f)/\|\phi(x, \tiltheta_f)\|_2$, and context vectors $\bar{\psi}(c, \tiltheta_c) = \psi(c, \tiltheta_c)/\|\psi(c, \tiltheta_c)\|_2$. When computing predictive variances at inference time, we rescale by 
\begin{align}
\mathbb{V}[f(x,\theta_\infty)] \approx \|\varphi(x, \theta_0^{1:L-1}) \|_2^2 \big( 
\bar{\varphi}(x, \theta_0^{1:L-1})^\top \bar{\varphi}(x, \theta_0^{1:L-1}) - \bar{\phi}(x, \tiltheta_f)^\top \bar{\psi}(c, \tiltheta_c) \big) \,,
\end{align}
to obtain predictions in the original scale again. 
\paragraph{Small function initialization.} While our theoretical suggests using small function initialization with $g(x,\tiltheta_0) \approx 0,\, \forall x$, preliminary experiments with a reparametrization $\hat{g}(x,\tiltheta_t) := g(x,\tiltheta_t) - g(x,\tiltheta_0)$ showed no significant differences. In our main implementation we thus refrain from using this reparametrization in favor of simplicity. 

 \begin{table}[p]
  \caption{Data augmentations for context data.}
  \centering
  \begin{tabular}{ll}
    \toprule
    Distribution Shift & VizDoom\\
    \midrule
    RandomHorizontalFlip($p=0.25$) & RandomPerspective($p=0.5$) \\
    RandomVerticallFlip($p=0.25$) & RandomHorizontalFlip($p=0.5$) \\
    Rotate($p=0.25$) & RandomResizedCrop($\text{r}=[0.75,1.0]$) \\
    GaussianBlur($\sigma=1.0, \, p=0.25$) & \\
    RandomResizedCrop($\text{r}=[0.75,1.0]$) &  \\
    RandomBrightness($\text{r}=[-1.0,1.0],\, p=0.5$) & \\
    RandomContrast($\text{r}=[-1.0,1.0],\, p=0.5$) & \\
    \bottomrule
  \end{tabular}
  \label{tab:dataaug}
  \vspace{-2mm}
\end{table}

\subsection{Additional Experimental Results} \label{sup:addexps}
\begin{figure}[h]
    \centering
        \includegraphics[width=0.65\columnwidth]{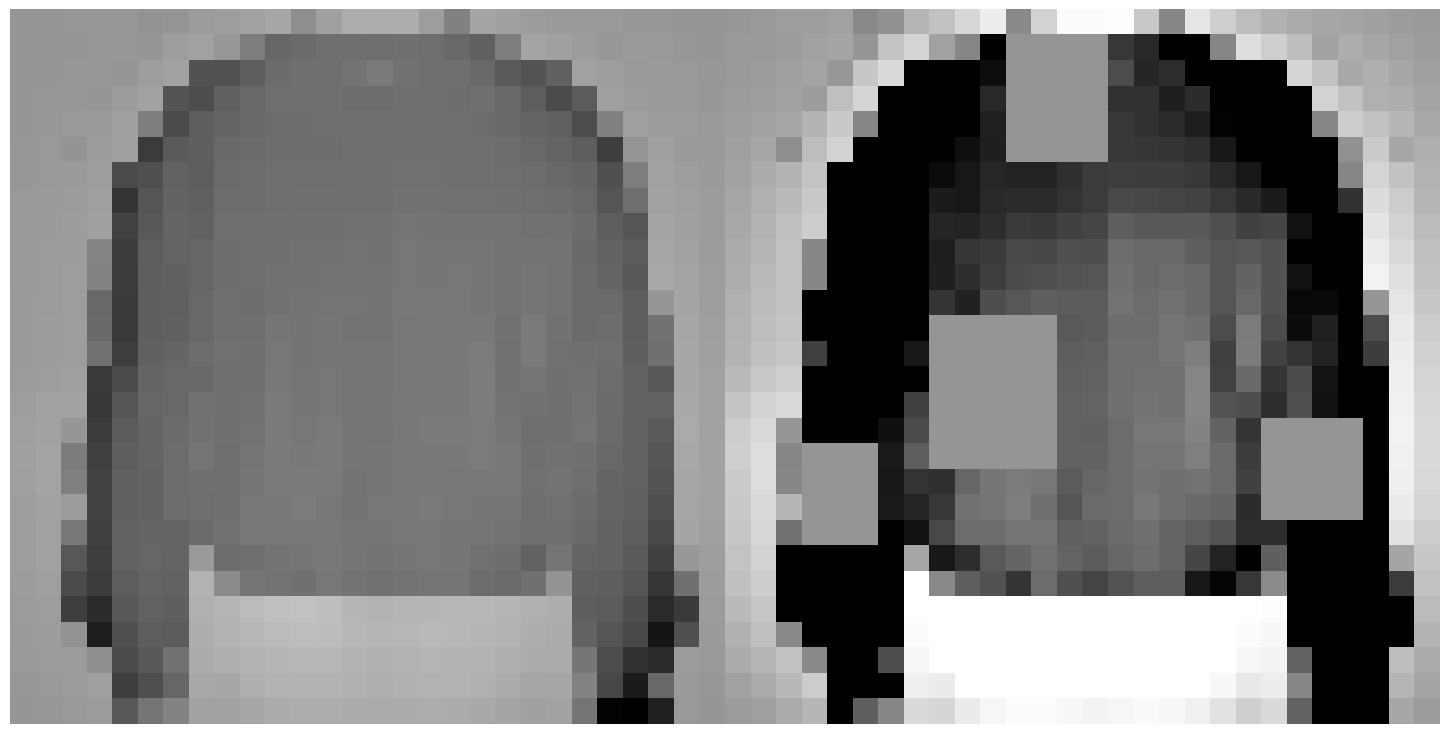}
        \caption{Left: Original Image. Right: Perturbed OOD Image.}
        \label{fig:ood_perturbed}
\end{figure}

We report the detailed results of our distribution shift detection tasks. Tables~\ref{tab:ooddetfm} to~\ref{tab:ooddetkm} show OOD detection metrics for the datasets FashionMNIST, MNIST, NotMNIST, and KMNIST. 
Each table shows the test accuracy and average AUROC, AUPR-IN and AUPR-OUT scores against the remaining three training datasets and an additional perturbed dataset. The perturbed dataset is constructed by applying data augmentations to the ID dataset. In our experiments, we use random brightness changes $(p=1.0, r=[-1.0,1.0])$, random contrast changes$(p=1.0, r=[-1.0,1.0])$, and randomly set patches of an image to zero $(p=1.0, r=[-1.0,1.0])$. Fig.~\ref{fig:ood_perturbed} shows an example of this. 

\begin{table}[p]
  \caption{Distribution Shift Detection. FashionMNIST as ID dataset.}
  \label{tab:ooddetfm}
  \centering
    \begin{tabular}{lllll}
        \toprule
        \textbf{Method} & \textbf{Acc.} & \textbf{AUROC} & \textbf{AUPR-IN} & \textbf{AUPR-OUT} \\
        \midrule
        MCD & $89.24\pm 0.36$ & $82.23\pm 0.48$ & $79.88\pm 0.75$ & $83.01\pm 0.34$ \\
        BNN-MCMC & $85.73\pm 0.24$ & $85.01\pm 0.62$ & $85.16\pm 0.68$ & $83.38\pm 0.62$ \\
        BNN-Laplace & $88.57\pm 0.80$ & $86.50\pm 0.67$ & $86.32\pm 0.75$ & $85.95\pm 0.75$ \\
        RND & $91.90\pm 0.15$ & $\bm{93.93}\pm 0.73$ & $\bm{93.45}\pm 1.12$ & $\bm{93.64}\pm 0.52$ \\
        ENS(3) & $92.90\pm 0.09$ & $88.90\pm 0.20$ & $89.63\pm 0.19$ & $88.16\pm 0.20$ \\
        ENS(15) & $\bm{93.33}\pm 0.06$ & $91.93\pm 0.12$ & $92.83\pm 0.11$ & $91.09\pm 0.12$ \\
        \midrule
        CSD & $91.93\pm 0.17$ & $\bm{96.18}\pm 0.67$ & $\bm{96.49}\pm 0.74$ & $\bm{95.74}\pm 0.62$ \\
        CSD-Aug. & $91.92\pm 0.16$ & $\bm{97.84}\pm 0.30$ & $\bm{98.24}\pm 0.27$ & $\bm{97.34}\pm 0.31$ \\
        CSD-OOD. & $91.96\pm 0.13$ & $\bm{97.35}\pm 0.50$ & $\bm{97.87}\pm 0.45$ & $\bm{96.72}\pm 0.56$ \\
        \bottomrule
    \end{tabular}
\end{table}

\begin{table}[p]
  \caption{Distribution Shift Detection. MNIST as ID dataset.}
  \label{tab:ooddetmn}
  \centering
    \begin{tabular}{lllll}
        \toprule
        \textbf{Method} & \textbf{Acc.} & \textbf{AUROC} & \textbf{AUPR-IN} & \textbf{AUPR-OUT} \\
        \midrule
        MCD & $98.97\pm 0.06$ & $90.03\pm 0.23$ & $87.70\pm 0.38$ & $89.01\pm 0.32$ \\
        BNN-MCMC & $94.29\pm 0.39$ & $80.24\pm 2.19$ & $80.20\pm 2.05$ & $77.33\pm 2.56$ \\
        BNN-Laplace & $94.17\pm 1.01$ & $74.05\pm 1.70$ & $72.24\pm 1.90$ & $74.39\pm 1.73$ \\
        RND & $99.85\pm 0.02$ & $94.66\pm 0.52$ & $93.83\pm 0.95$ & $\bm{94.25}\pm 0.35$ \\
        ENS(3) & $99.95\pm 0.01$ & $94.03\pm 0.24$ & $95.09\pm 0.22$ & $92.32\pm 0.31$ \\
        ENS(15) & $\bm{99.97}\pm 0.00$ & $\bm{95.33}\pm 0.06$ & $\bm{96.31}\pm 0.06$ & $93.79\pm 0.10$ \\
        \midrule
        CSD & $99.88\pm 0.01$ & $\bm{96.78}\pm 0.58$ & $\bm{96.96}\pm 0.72$ & $\bm{96.25}\pm 0.57$ \\
        CSD-Aug. & $99.87\pm 0.02$ & $\bm{98.39}\pm 0.17$ & $\bm{98.63}\pm 0.20$ & $\bm{97.94}\pm 0.19$ \\
        CSD-OOD. & $99.87\pm 0.02$ & $\bm{99.37}\pm 0.08$ & $\bm{99.51}\pm 0.07$ & $\bm{99.14}\pm 0.11$ \\
        \bottomrule
    \end{tabular}
\end{table}

\begin{table}[p]
  \caption{Distribution Shift Detection. NotMNIST as ID dataset.}
  \label{tab:ooddetnm}
  \centering
    \begin{tabular}{lllll}
        \toprule
        \textbf{Method} & \textbf{Acc.} & \textbf{AUROC} & \textbf{AUPR-IN} & \textbf{AUPR-OUT} \\
        \midrule
        MCD & $95.17\pm 0.14$ & $83.21\pm 0.45$ & $75.86\pm 0.89$ & $85.73\pm 0.18$ \\
        BNN-MCMC & $90.20\pm 0.44$ & $87.05\pm 0.80$ & $85.93\pm 1.10$ & $87.68\pm 0.63$ \\
        BNN-Laplace & $95.29\pm 0.52$ & $86.38\pm 1.46$ & $82.99\pm 2.36$ & $87.55\pm 1.04$ \\
        RND & $96.25\pm 0.12$ & $\bm{95.49}\pm 0.82$ & $\bm{95.81}\pm 0.97$ & $\bm{95.23}\pm 0.74$ \\
        ENS(3) & $97.12\pm 0.08$ & $92.37\pm 0.26$ & $92.11\pm 0.30$ & $91.93\pm 0.27$ \\
        ENS(15) & $\bm{97.47}\pm 0.05$ & $94.04\pm 0.16$ & $94.26\pm 0.17$ & $93.29\pm 0.17$ \\
        \midrule
        CSD & $96.48\pm 0.08$ & $\bm{96.98}\pm 0.41$ & $\bm{97.26}\pm 0.44$ & $\bm{96.86}\pm 0.36$ \\
        CSD-Aug. & $96.45\pm 0.09$ & $\bm{98.51}\pm 0.22$ & $\bm{98.70}\pm 0.24$ & $\bm{98.31}\pm 0.21$ \\
        CSD-OOD. & $96.49\pm 0.10$ & $\bm{98.49}\pm 0.35$ & $\bm{98.78}\pm 0.29$ & $\bm{98.21}\pm 0.42$ \\
        \bottomrule
    \end{tabular}
\end{table}

\begin{table}[p]
  \caption{Distribution Shift Detection. KMNIST as ID dataset.}
  \label{tab:ooddetkm}
  \centering
    \begin{tabular}{lllll}
        \toprule
        \textbf{Method} & \textbf{Acc.} & \textbf{AUROC} & \textbf{AUPR-IN} & \textbf{AUPR-OUT} \\
        \midrule
        MCD & $94.18\pm 0.26$ & $87.22\pm 0.75$ & $83.48\pm 0.74$ & $88.00\pm 0.77$ \\
        BNN-MCMC & $80.57\pm 1.29$ & $80.40\pm 1.46$ & $79.31\pm 1.93$ & $80.75\pm 1.31$ \\
        BNN-Laplace & $85.39\pm 1.79$ & $78.58\pm 2.66$ & $76.18\pm 3.11$ & $79.47\pm 2.49$ \\
        RND & $96.73\pm 0.21$ & $93.50\pm 1.17$ & $93.58\pm 1.45$ & $92.93\pm 1.05$ \\
        ENS(3) & $97.68\pm 0.10$ & $93.88\pm 0.24$ & $94.49\pm 0.26$ & $93.05\pm 0.24$ \\
        ENS(15) & $\bm{97.96}\pm 0.06$ & $\bm{94.68}\pm 0.11$ & $\bm{95.39}\pm 0.12$ & $\bm{93.81}\pm 0.11$ \\
        \midrule
        CSD & $96.89\pm 0.13$ & $\bm{96.57}\pm 0.73$ & $\bm{97.05}\pm 0.74$ & $\bm{95.90}\pm 0.74$ \\
        CSD-Aug. & $96.90\pm 0.19$ & $\bm{98.12}\pm 0.46$ & $\bm{98.45}\pm 0.41$ & $\bm{97.61}\pm 0.53$ \\
        CSD-OOD. & $96.86\pm 0.12$ & $\bm{99.06}\pm 0.19$ & $\bm{99.30}\pm 0.14$ & $\bm{98.71}\pm 0.25$ \\
        \bottomrule
    \end{tabular}
\end{table}

\end{document}